\pdfoutput=1

\documentclass[11pt]{article}

\usepackage{acl}

\usepackage{times}
\usepackage{latexsym}

\usepackage[T1]{fontenc}

\usepackage[utf8]{inputenc} 

\usepackage{microtype}

\usepackage{inconsolata}

\usepackage{graphicx}
\usepackage{booktabs}
\usepackage{amsmath}
\usepackage{float}
\usepackage{tikz}

\title{Automated Essay Scoring Incorporating Annotations from Automated Feedback Systems}

\author{Christopher Ormerod \\
  Cambium Assessment Inc. \\ 
  {\tt christopher.ormerod@cambiumassessment.com}}

\date{}

\begin{document}

\maketitle

\begin{abstract}
This study illustrates how incorporating feedback-oriented annotations into the scoring pipeline can enhance the accuracy of automated essay scoring (AES). This approach is demonstrated with the Persuasive Essays for Rating, Selecting, and Understanding Argumentative and Discourse Elements (PERSUADE) corpus. We integrate two types of feedback-driven annotations: those that identify spelling and grammatical errors, and those that highlight argumentative components. To illustrate how this method could be applied in real-world scenarios, we employ two LLMs to generate annotations -- a generative language model used for spell correction and an encoder-based token-classifier trained to identify and mark argumentative elements. By incorporating annotations into the scoring process, we demonstrate improvements in performance using encoder-based large language models fine-tuned as classifiers.
\end{abstract}


\section{Introduction}

Automated Essay Scoring (AES) uses statistical models to assign grades to essays that approximate hand-scoring \cite{shermis_contrasting_2013}. Automated Writing Evaluation (AWE) is the provision of automated feedback designed to help students iteratively improve their essays \cite{huawei_systematic_2023}. Initial attempts at AES and AWE were based on Bag-of-Words (BoW) models that combine frequency-based and hand-crafted features \cite{attali_automated_2006, page_project_2003}. Well-designed features can serve two purposes: to improve scoring accuracy and provide feedback to students to improve their essays. These features tend to be global features, such as the number of words, sentence length, or readability metrics, and are not based on fine-grained semantics or the organizational structure of essays.

Many modern AES engines employ transformer-based Large Language Models (LLM)s \cite{rodriguez_language_2019}, which offer improved accuracy over bag-of-words models. However, this comes at the expense of reduced interpretability due to implicit feature definition. Some researchers have sought to combine LLM-derived features with traditional hand-crafted features to enhance accuracy and provide some level of interpretability \cite{uto_automated_2020}. LLMs also offer the ability to provide semantically rich feedback, such as key phrases from explainable AI \cite{boulanger_shaped_2020}, annotation schemas for automated writing evaluation systems \cite{crossley_persuasive_2022, lottridge_write_2024}, and the generation of detailed feedback through LLM prompting techniques \cite{lee_applying_2024}. This study considers how these semantically rich features, designed primarily for providing feedback, can also enhance scoring accuracy.

We demonstrate our approach using the Persuasive Essays for Rating, Selecting, and Understanding Argumentative and Discourse Elements (PERSUADE) corpus, which is a dataset of essays in which the argumentative components of the essays were annotated \cite{crossley_persuasive_2022}. This dataset also contains scores assigned against an openly available holistic rubric \footnote{https://github.com/scrosseye/persuade\_corpus\_2.0} and demographic data, which allows us to test any AES system with respect to operational standards, including the addition of potential bias \cite{williamson_framework_2012}. 

The two classes of features we consider are derived from Grammatical Error Correction (GEC) and Computational Argumentation. The goal of GEC is to provide a mapping from a sentence that may or may not contain errors in language, to a version with the same meaning with fewer language errors \cite{martynov_methodology_2023}, while computational argumentation seeks to isolate and analyze the set of argumentative components of an essay \cite{stab_annotating_2014}. For these features to be incorporated into an AES pipeline, we leverage the ability of language models to accurately annotate argumentative components \cite{ormerod_argumentation_2023} and correct spelling and grammatical errors \cite{rothe_simple_2021}. These spelling and grammatical corrections can then be annotated and classified to provide locally defined information on conventions \cite{korre_errant_2020}. Once these annotations have been derived, we incorporate the annotations using Extensible Markup Language (XML) for easy parsing, which facilitates natural integration into an AWE system based on essays encoded in HTML.

We must also be cognizant that increased automation can exacerbate biases, especially for English Language Learners \cite{ormerod_short-answer_2022}. For this reason, we also examine whether this pipeline leads to greater bias by examining the standardized mean difference for the relevant subgroups. 

In \S \ref{sec:method}, we describe our methods, including the data used, the models, and the approach. The performance of the annotation models and the scoring models are presented in \S \ref{sec:results}. We will discuss the findings and suggest future directions in \S \ref{sec:discussion}.

\section{Method}\label{sec:method}

\subsection{Data}

The PERSUADE corpus is an openly available dataset of 25,996 argumentative essays between grades 6 and 12 on a range of 15 different topics \cite{crossley_persuasive_2022}. The essays are responses to prompts that are either dependent on source material, or independent of source material. The set has been divided into a training set and a test set by the original authors. An outline of the composition of these two sets is presented in Table \ref{tab:train-test}. 

\begin{table}
\begin{center}
\begin{tabular}{l r r r | r r  r} \toprule
& & & & {\bf Avg.} \\
{\bf Grade} & {\bf Train} & {\bf Test} & {\bf Total} & {\bf Len.} \\ \midrule
6 & 688   & 684  & 1372 & 294.6 \\
8 & 5614  & 4015 & 9629 & 374.9\\
9 & 1831  & 235  & 2066 & 426.6 \\
10 & 4654 & 3620 & 8274 & 407.6\\
11 & 1863 & 1220 & 3083 & 610.9\\
12 & 243  & 161  & 404  & 469.0\\
Unk. & 701 & 467 & 1168 & 452.5 \\ \midrule
Total & 15594 & 10402 & 25996 & 418.1 \\ \bottomrule
\end{tabular}
\end{center}
\caption{The grade level and length statistics for the training and testing splits for the PERSUADE corpus, and the counts of essays that are responses to prompts that are dependent (Dep.) on source material and independent (Ind..) of source material. \label{tab:train-test}}
\end{table}

\subsubsection{Annotations}

A key characteristic of the corpus that makes it useful from the standpoint of computational argumentation is the annotations. The argumentative clauses of each essay were identified and classified into one of seven classes:

\begin{enumerate}
    \item {\bf Lead (L):} An introduction that begins with a statistic, a quotation, a description, or some other device to grab the reader’s attention and point toward the thesis.
    \item {\bf Position (P):} An opinion or conclusion on the main question
    \item {\bf Claim (C1)}: A claim that supports the position
    \item {\bf Counterclaim (C2)}: A claim that refutes another claim or gives an opposing reason to the position
    \item {\bf Rebuttal (R)}: A claim that refutes a counterclaim
    \item {\bf Evidence (E)}:  Ideas or examples that support claims, counterclaims, rebuttals, or the position
    \item {\bf Concluding Statement (C3)}: A concluding statement that restates the position and claims
\end{enumerate}

There are an average of 11.0 annotated components in each essay. Some descriptive statistics on the distribution of applied labels can be found in Table \ref{tab:labels}. Among the labels, the most frequently applied labels are Claims and Evidence and the least frequently applied labels are Counterclaims and Rebuttals. In accordance with state standards, the development of a counter-argument in persuasive essay writing is developed at grades eight and beyond, hence, Counterclaims and Rebuttals are rarely applied at the sixth-grade level. 

\begin{table}[H]
\begin{center}
\begin{tabular}{l r r r r r r} \toprule
& {\bf 6} & {\bf 8} & {\bf 9} & {\bf 10} & {\bf 11} & {\bf 12} \\ \midrule
L        & 4.3 & 4.5 & 4.7 & 5.9 & 6.0 & 5.0  \\
P     & 9.8 & 9.0 & 9.1 & 9.9 & 7.0 & 8.3 \\
C1       & 27.2 & 27.7 & 27.6 & 29.8 & 31.2 & 34.3  \\
C2     & 2.1 & 2.8 & 5.2 & 2.5 & 5.3 & 5.0 \\
R     & 1.5 & 2.2 & 3.7 & 1.7 & 4.3 & 2.0  \\
E    & 27.3 & 25.7 & 26.8 & 28.7 & 23.4 & 27.5  \\
C3       & 8.0 & 7.5 & 7.6 & 8.7 & 6.9 & 7.5 \\ \bottomrule
\end{tabular}
\end{center}
\caption{Some descriptive statistics regarding the distribution of annotations with respect to the various grade levels. \label{tab:labels}}
\end{table}

We did not use the effectiveness scores for the discourse elements in this study. This was a conscious choice due to the fairly low agreement between the effectiveness scores assigned by human raters ($\kappa = 0.316$). Similar attempts at judging the quality of arguments have also yielded low agreement rates \cite{gretz_large-scale_2019, toledo_automatic_2019}.

\subsubsection{Scores}

Each essay was graded against a standardized SAT holistic essay scoring rubric, which was slightly modified for the source-based essays \footnote{https://www.kaggle.com/datasets/davidspencer/persuade-rubric-holistic-essay-scoring}. Based on the rubric, a high-scoring essay (5-6) demonstrates mastery through effective development of a clear point of view, strong critical thinking with appropriate supporting evidence, well-organized structure with coherent progression of ideas, skillful language use with varied vocabulary and sentence structure, and minimal grammatical errors. In contrast, a lower-scoring essay (1-3) exhibits significant weaknesses: vague or limited viewpoint, weak critical thinking with insufficient evidence, poor organization resulting in disjointed presentation, limited vocabulary with incorrect word choices, frequent sentence structure problems, and numerous grammatical errors that interfere with meaning. 

The score distribution is fairly regular, with the highest and lowest scores being the rarest. The full score distribution can be found in Table \ref{tab:score_dist}. The reported inter-rated reliability, as reported in \cite{crossley_persuasive_2022}, is given by $\kappa = 0.745$ (see \eqref{eq:qwk}). 

\begin{table}[H]
    \centering
    \begin{tabular}{l|r r r r r r} \toprule
        Score & 1 & 2 & 3 & 4 & 5 & 6 \\ \midrule
        \% & 4.0 & 21.9 & 32.2 & 25.9 & 12.7 & 3.4\\ \bottomrule
    \end{tabular}
    \caption{The score distribution for the PERSUADE dataset.\label{tab:score_dist}}
\end{table}

The key differentiators in the rubric between high and low scoring essays are the clarity of thought, quality of supporting evidence, organizational coherence, and technical proficiency in language use. The premise behind the approach is that organizational coherence and technical proficiency in language are both made clearer by highlighting the argumentative components and conventions-based errors. Provided our pipeline for annotating essays is sufficiently accurate, these annotations should help the engine align scores with the rubric. 

\subsubsection{Augmented Data}

The input into the scoring model was augmented to use the annotation information using Extensible Markup Language (XML). We have an XML tag per argumentative component type. To annotate conventions errors, we used the ERRANT tool \cite{bryant_automatic_2017}. The ERRANT tool classifies errors into 25 different main types, with many of these categories appearing with three different subtypes: "R" for replace, "M" for missing, and "U" for Unnecessary. For example, one category, "PUNCT", refers to a punctuation error. We can either replace, remove, or add punctuation to a sentence to make it correct, corresponding to "R:PUNCT", "U:PUNCT", and "M:PUNCT", respectively. We refer to \cite{bryant_automatic_2017} for a full explanation of the categories. 

To simplify the categories for annotation purposes, we divide all possible ERRANT annotations into three labels: $\verb\<Spelling\>$, $\verb\<PunctOrth\>$, and $\verb\<Grammar\>$. The $\verb\<Spelling\>$ label is applied to the subcategories of "SPELL", the $\verb\<PunctOrth\>$ is applied to the subcategories of "PUNCT" and "ORTH", while all other categories are designated as having labels of $\verb\<Grammar\>$. This means that we have a total of 10 annotation labels: 7 associated with argumentative components and 3 for convention errors. An example of the input into the model is shown in Figure \ref{fig:example-input}. Since we do not have human-annotated data, the augmented data relies on the output from an annotation model and a spell-correction model, both of which can contribute to annotations that are less accurate. 

\begin{figure}[H]
\begin{center}
\begin{verbatim}
<Lead>There can be many <Spelling>
advanteges</Spelling> and <Spelling>
disadvanteges</Spelling> to having a
car but the <Spelling>advanteges
<Spelling> to not having a <Grammar>card
</Grammar> greatly outweighs having one. 
</Lead>There can be many reasons why not
having a car is great but the main three
are <Claim>it reduces pollution reduces
stress</Claim> and <Claim>having less
cars reduces the <Spelling>noice 
<Spelling> pollution in a city. </Claim>
Can you imagine a place with no cars?
\end{verbatim}
\end{center}
\caption{An example of the model input using an excerpt of an annotated essay. \label{fig:example-input}}
\end{figure}

{\bf Demographics}: The last detail of this dataset, which makes it exceptionally well-suited to the investigation of AES in an operational setting, is the accompanying demographic information. This allows us to investigate any additional potential bias introduced in modeling. We measure bias by the original operational standards defined by Williamson et al. \cite{williamson_framework_2012}. While this standard is important for many reasons, there have been numerous alternative approaches to bias \cite{ormerod_automated_2022}. 

\begin{table}[H]
\begin{center}
\begin{tabular}{l l r r} \toprule
key & Subgroup & Train & Test \\ \midrule
WC & White/Caucasian & 7012 & 4559\\
HL &Hispanic/Latino & 3869 & 2691\\
BA & Black/African  & 2975 & 1984\\
& American\\
AP & Asian/Pacific Islander & 1072 & 671 \\
Mix & Two or more & 598 & 424\\
Nat & American Indian & 68 & 73 \\ &Alaskan Native\\ \midrule
ELL & English Language & 1330 &  914\\ & Learner  \\
DE & Disadvantaged  & 5391 & 4252 \\ & Economically\\
ID & Identified Disability & 1516 & 1172 \\ \bottomrule
\end{tabular}
\end{center}
\caption{The main subgroup populations in the train and test set. \label{tab:subgroups}}
\end{table}

The population of various subgroups in the train and test split, as presented in the data, have been outlined in Table \ref{tab:subgroups}. 

\subsection{Modeling details}

Since the advantages of the transformer were first celebrated \cite{vaswani_attention_2017} and BERT was trained \cite{devlin_bert_2018}, many of the state-of-the-art results can be attributed to transformer-based LLMs \cite{wang_glue_2019}. For this reason, we restrict our attention to fine-tuned transformer-based LLMs, whose architectures can be described as encoder, decoder, or encoder-decoder models \cite{vaswani_attention_2017}. As a general rule, encoder models excel in natural language inference tasks \cite{devlin_bert_2018}, decoder models excel in generative tasks \cite{radford_improving_2018}, and encoder-decoder models excel in translation, where the task benefits from representation learning \cite{raffel_exploring_2020}.

\subsubsection{Annotation model} 

The task of annotations can be framed as a token-classification task, where each token is classified into one of eight possible labels, one for each possible argumentative component in addition to one extra label for unannotated regions. This task lends itself to an encoder-based model trained as a masked language model like BERT \cite{devlin_bert_2018}. Since BERT, arguably the best performing series of models are Microsofts' DeBERTa model series \cite{he_deberta_2021}. The problem with these models is that many of the essays exceed the 512 token limit after tokenization. For this reason, we turn to a newly developed long context model known as ModernBERT \cite{warner_smarter_2024}. 

Aside from some differences in the choices of normalization layers \cite{xiong_layer_2020}, the use of gated activation functions \cite{shazeer_glu_2020}, and more extensive pretaining, the biggest difference in the architecture is the use of Rotational Positional Embeddings (RoPE) \cite{su_roformer_2024}. To understand how RoPE works, in the original implementation of attention, the output of attention is given as a function of the key vectors, $k_i$, query vectors, $q_j$, and value vectors, $v_l$, given by  
\begin{equation}
a_{m,n} = \frac{\exp (q_m^T k_n/ \sqrt{d}) }{\sum_j \exp (q_m^T k_j/ \sqrt{d})}.
\end{equation}
where key, query, and value vectors are functions of the embedding vectors at the first attention layer. The standard construction is that these functions be affine linear functions (linear with a bias term), where the positional embedding is the addition of token embeddings and some learnable positional embedding terms. Instead of adding a vector, we define the query and key vectors using 
\begin{equation}
q_m = R_{\Theta,m}^{d} W_q x_m, \hspace{1cm} k_m = R_{\Theta,m}^{d} W_k x_m
\end{equation}
where $R_{\Theta,m}^{d}$ a block-diagonal matrix of 2-dimensional rotation matrices, $r_\phi$, given by
\begin{align}
R_{\Theta,m}^{d} &= \mathrm{diag}( r_{m\theta_1}, \ldots, r_{m\theta_{d/2}}),\\
r_\phi &= \begin{pmatrix} \cos \phi & -\sin \phi \\ \sin\phi & \cos\phi \end{pmatrix} ,
\end{align}
with $\theta_i = \tilde{\vartheta}^{-2(i-1)/d}$. The term $\vartheta$ is called the RoPE Theta. The key idea is that this transformation encodes positional information by rotating token embeddings in a particular manner that allows the model to understand relative distances between tokens rather than just absolute positions. 

The ModernBERT model, like BERT, is trained as a masked-language model with an encoder and a token classification head. The structure of the encoder includes multiple layers of self-attention in which the attention layers alternate between rotary embeddings with base $\vartheta = 1 \times 10^4$ and $\vartheta = 1.6\times 10^5$. However, the training was done in two phases. By adjusting the value of $\vartheta$, researchers developed a way to scale the context length of the RoPE embedding \cite{fu_data_2024}, which has been applied to many other models \cite{aimeta_llama_2024}. Using this technique, the ModernBERT model was pretrained on 1.7 trillion tokens with a context length of 1024 with a $\vartheta= 10^{-4}$, which was extended to 8196 by additional training with altered layers in which $\vartheta = 1.6\times 10^5$. In this manner, one way of thinking of the change in $\vartheta$ values in the encoder is that the attention mechanism alternates between global and local attention.

It should be noted that many architectures have attempted to circumvent the context limitation, such as the Reformer \cite{kitaev_reformer_2020}, Longformer \cite{beltagy_longformer_2020}, Transformer-XL \cite{dai_transformer-xl_2019}, and XLNet \cite{yang_xlnet_2019}, to name a few. Many of these solutions use some sort of sliding context window and/or a recurrent adaptation of the transformer architecture. Extending the context using rotary positional embeddings is more computationally efficient and effective at scaling to large context lengths, making ModernBERT a more appropriate choice in this context. 

The pretrained ModernBERT model was modified for annotation by adding a classification head to the encoder. The annotator model possesses a classification head with 9 output dimensions: 7 for argumentative component labels, 1 for unannotated text, and 1 for padded variables (which can be disregarded). The output represents log probabilities for each label. This model was trained to predict the annotations for each token in the training set using the cross-entropy loss function and the Adam optimizer with a learning rate of 1e-6 over 10 epochs. We simplified the training by not having a development set.

\subsubsection{Spelling and Grammar}

The most successful and accurate way to perform GEC has been to utilize representation learning, hence, we seek an encoder-decoder model, which is a sequence-to-sequence model \cite{sutskever_sequence_2014}. The premise behind this method is that we are able to use the encoder to map sentences to a vector space that encodes the semantic information, while a decoder maps from the vector space to grammatically correct text \cite{rothe_simple_2021}. This suggests we use a T5 model, pretrained as a text-to-text transformer and fine-tuned to perform grammatical error correction (GEC) \cite{martynov_methodology_2023}. Once the correction is defined, the original sentence and the correction are used as input into the ERRANT tool to produce a classified correction \cite{korre_errant_2020}. 

\subsubsection{Scoring Models}

Given the model input includes both the essay and any annotations, we require long-context models. While we have experimented with the use of QLoRA-trained generative Models \cite{ormerod_automated_2024}, to simplify the presentation, we use the ModernBERT model for scoring in addition to annotation. In this case, the scoring models were constructed by appending a linear classification head to the ModernBERT model with 6 targets, one for each score point.

\begin{figure}
\begin{tikzpicture}[yscale=0.9, xscale=1.3]
\node[draw=black, very thick, rounded corners=3pt, fill = green!10, minimum height=3em, minimum width=6em](orig) at (0,4) {Essay};
\node[draw=black, very thick, rounded corners=3pt, fill = blue!10](bert) at (-2,2) {\begin{tabular}{c}ModerBERT\\Annotator \end{tabular}};
\node[draw=black, very thick, rounded corners=3pt, fill = blue!10](score) at (0,-4) {\begin{tabular}{c}ModerBERT\\Scorer \end{tabular}};
\node[draw=black, very thick, rounded corners=3pt, fill = green!10](ann1) at (-2,-2) {\begin{tabular}{c}Argument\\ Annotated\\ Essay \end{tabular}};
\node[draw=black, very thick, rounded corners=3pt, fill = green!10](ann2) at (0,-2) {\begin{tabular}{c}Combined\\Annotated\\ Essay \end{tabular}};
\node[draw=black, very thick, rounded corners=3pt, fill = green!10](ann3) at (2,-2) {\begin{tabular}{c}Error\\Annotated\\ Essay \end{tabular}};
\node[draw=black, very thick, rounded corners=3pt, fill = blue!10](errant) at (2,0) {\begin{tabular}{c}ERRANT \end{tabular}};
\node[draw=black, very thick, rounded corners=3pt, fill = blue!10](gec) at (2,2) {\begin{tabular}{c}T5-GEC\\Model\end{tabular}};
\draw[rounded corners = 10pt,->,thick] (orig) -| (gec);
\draw[rounded corners = 10pt,->, thick] (orig) |- (errant);
\draw[rounded corners = 10pt,->, thick] (gec) -- (errant);
\draw[rounded corners = 10pt,->, thick] (errant) -- (ann3);
\draw[thick, ->](bert) -- (ann1);
\draw[thick, ->,rounded corners = 10pt,](orig) -| (bert);
\draw[thick, ->](ann2) -- (score);
\draw[thick, ->](ann1) -- (ann2);
\draw[thick, ->](ann3) -- (ann2);
\end{tikzpicture}
\caption{A diagram representing the scoring pipeline.\label{fig:pipeline}}
\end{figure}

\subsection{Evaluation}

We have two models to evaluate: an annotation model and a classification model. There are many challenges to assessing annotations for argumentative clauses. We need to carefully define what it means for a particular clause to be identified and correctly classified, given that certain identifications may not perfectly align with the predicted components. For holistic scoring, there are many more well-defined and accepted standards presented by Williamson et al. \cite{williamson_framework_2012}.

\subsubsection{Annotator Evaluations}

The standard described by the annotated argumentative essay dataset \cite{stab_annotating_2014, stab_identifying_2014} is reduced to a classification of IOB tags, where the governing statistic is an F1 score. Our guiding principle is the original rules for the competition\footnote{https://www.kaggle.com/c/feedback-prize-2021/overview}, in which we use the ground truth and consider a match if there is over 50\% overlap between the two identified components. Given a match using this rule, matching the argumentative type is considered a true positive (TP), unmatched components are considered false negatives (FN), while predicted label mismatches are considered false positives (FP). The final reported value is the F1 score, given by the familiar formula
\begin{equation}\label{eq:f1}
F1 = \frac{2TP}{2TP + FP + FN}.
\end{equation}
We can compare agreements for each label applied based on the ground truth. In this way, for each type of argumentative component type, we have a corresponding F1 score. In accordance with the rules of the competition, the final F1 statistic of interest is the macro average, given by the unweighted average over all the classes. 

\subsubsection{Error Annotations}

When it comes to annotating errors in the use of language, since the errors in the essays were not explicitly annotated by hand, we have no direct way of evaluating the accuracy of any annotations. We can only rely on the accuracy of the individual components. The T5 model we used has been evaluated in \cite{martynov_methodology_2023} with respect to the JFLEG dataset \cite{napoles_jfleg_2017} and the BEA60k dataset \cite{jayanthi_neuspell_2020}. According to those benchmarks, the accuracy of the model used is comparable to ChatGPT and GPT-4. 

\subsubsection{Automated Scoring Evaluations}

For the scoring model, we use the standards for agreement specified by Williamson et al. \cite{williamson_framework_2012}. The first and primary statistic used to describe agreement is Cohen's quadratic weighted kappa (QWK) \cite{cohen_coefficient_1960}. Given scores between $1$ and $N$, we define the weighted kappa statistic by the formula
\begin{equation}\label{eq:qwk}
\kappa = 1 - \frac{\sum w_{ij} O_{ij}}{\sum w_{ij} E_{ij}}
\end{equation}
where $O_{ij}$ is the number of observed instances where the first rater assigns a score of $i$ and the second rater assigns a score of $j$, and $E_{ij}$ are the expected number of instances that first rater assigns a score of $i$ and the second rater assigns a score of $j$ based purely on the random assignment of scores given the two rater's score distribution. This becomes the QWK when we apply the quadratic weighting:
\begin{equation}\label{eq:weight}
w_{ij} = \dfrac{(i-j)^2}{(N-1)^2}.
\end{equation}
The QWK takes values from $-1$ and $1$, indicating perfect disagreement and agreement, respectively. It is often interpreted as the probability of agreement beyond random chance. The second statistic used is exact agreement, which is viewed as less reliable since uneven score distributions can skew it. 

The last statistic used is the standardized mean difference (SMD). If $y_t$ represents the true score and $y_p$ represents the predicted score, then the SMD is given by 
\begin{equation}\label{eq:smd}
SMD(y_t, y_p) = \frac{\overline{y_p} - \overline{y_t}}{\sqrt{(\sigma(y_p)^2 + \sigma(y_t)^2)/2}}.
\end{equation}
This statistic can be interpreted as a standardized relative bias. A positive or negative value indicates that the model is introducing some positive  or negative bias in the modeling process, respectively. Furthermore, when we restrict this calculation to scores for a specific demographic, assuming that demographic is sufficiently well-represented, the SMD is considered a gauge of the bias associated with the modeling for that subgroup.

\section{Results}\label{sec:results}

\subsection{Annotation Accuracy}

Using the 50\% overlap rule, we present the number of true positives, false positives, and false negatives, and the resulting F1 score for each of the component types, excluding unannotated. These results are presented in Table \ref{tab:results_annotator}.  

\begin{table}[H]
    \centering
    \begin{tabular}{l|r r r r } \toprule
         &  TP & FP & FN & F1\\ \midrule
        L & 4332 & 270 & 95 & 0.960 \\
        P & 6359 & 832 & 205 & 0.924\\
        C1 & 20764 & 2057 & 1094 & 0.929\\
        C2 & 1839 & 654 & 164 & 0.818\\
        R & 1443 & 495 & 106 & 0.827\\
        E & 18838 & 1653 & 1299 & 0.927\\
        C3 & 5874 & 321 & 321 & 0.950\\ \bottomrule
    \end{tabular}
    \caption{The number of true positives (matched components and labels), false positives (matched components, unmatched labels), and false negatives (unmatched components) between the true annotations and the predicted annotations by component type.}
    \label{tab:results_annotator}
\end{table}

These scores are exceptionally high, with the lowest performance given by the annotator's ability to discern counterclaims and rebuttals. As an indication of the annotated errors in language, the pipeline highlighted 2795 spelling errors, 1401 grammatical errors, and 201 punctuation or orthography errors. The pipeline used does not seem to be uncovering as many errors as expected.


\subsection{Scoring Accuracy}

Given that the classification head is typically randomly initialized, we were also interested in whether these results were stable. We trained the scorer 10 times and reported the average, minimum, and maximum agreement levels for each agreement statistic we listed above. These statistics can be found in Table \ref{tab:results_scorer}.

\begin{table*}
    \centering
    \begin{tabular}{l|r r r | c c c | c c c} \toprule
                  &  \multicolumn{3}{c}{QWK} & \multicolumn{3}{c}{Exa} & \multicolumn{3}{c}{SMD} \\ 
                  & Avg & Min & Max & Avg & Min & Max & Avg & Min & Max \\  \midrule
                 Full Text Only       & 0.860 & 0.859 & 0.862 & 67.2 & 66.9 & 67.5 & -0.023 & -0.027 & -0.018\\
                 Component Annotated  & 0.868 & 0.867 & 0.870 & 68.9 & 68.6 & 69.1 & -0.013 & -0.017 & -0.009\\ 
                 Error Annotated      & 0.858 & 0.856 & 0.859 & 67.1 & 66.9 & 67.5 & -0.025 & -0.028 & -0.022\\
                 Combined Annotations & 0.866 & 0.867 & 0.870 & 68.4 & 68.0 & 68.9 &  0.021 & 0.016 & 0.025\\ 
                 Human Baseline       & 0.745 \\ \bottomrule
    \end{tabular}
    \caption{The average QWK, Exa, and SMD results on the test set for over 10 trials of training the scoring model.}
    \label{tab:results_scorer}
\end{table*}

All the models performed well above the human baseline. Out of the 10 separate trials, no model trained on the full-text alone scored as accurately as any of the models trained on the component annotated data or the data with both argumentative components and language errors annotated. An interesting observation is that the SMD, as calculated by \eqref{eq:smd}, is only positive for models trained on the combined annotations, and that the models trained on component annotated text showed the most controlled SMDs. 

\subsection{Potential bias}

To investigate the possibility of potential bias, we consider the SMD defined on subgroups. These SMDs are presented in Table \ref{tab:bias}. 

\begin{table}[H]
\begin{center}
\begin{tabular}{lrrrr}
\toprule 
Key & Orig. & Comp. & Error & Comb. \\ 
\midrule
Female & 0.12 & 0.11 & 0.12 & 0.06 \\ \midrule
WC & 0.09 & 0.11 & 0.08 & 0.15 \\
HL & -0.25 & -0.25 & -0.24 & -0.19 \\
BA & -0.17 & -0.16 & -0.20 & -0.17 \\
AP & 0.44 & 0.45 & 0.53 & 0.52 \\
Nat & -0.43 & -0.41 & -0.44 & -0.21 \\
Mix & 0.08 & 0.07 & 0.12 & 0.01 \\ \midrule
ELL & -0.59 & -0.60 & -0.62 & -0.54 \\
DE & -0.36 & -0.35 & -0.37 & -0.32 \\
ID & -0.51 & -0.48 & -0.49 & -0.42 \\
\bottomrule
\end{tabular}
\end{center}
\caption{The bias in the various subgroups as measured by SMD for the particular subgroup. \label{tab:bias}}
\end{table}

While the resulting bias was higher than expected, a cursory look seems to suggest that the use of combined annotations is mitigating some of the biases rather than exacerbating them. 

This work is one of a number of works that highlight the growing need to address the bias introduced by automation, especially for ELL students \cite{ormerod_automated_2022}. This suggests we need to apply bias mitigation, which could be of the form of a regression-based system with adjusted cut-off points \cite{ormerod_mapping_2022}, or some sort of reinforcement learning mechanism. 

\section{Discussion}\label{sec:discussion}

The results of this study demonstrate that incorporating feedback-oriented annotations into automated essay scoring (AES) pipelines can significantly improve scoring accuracy and provide meaningful, interpretable insights for students. The work of \citet{uto_automated_2020} suggests this is also true for traditional global features. What we propose is a realignment of AES to incorporate AWE elements so that we can provide students with more than just a score. 

One of the most critical findings from this study concerns the potential for introduced bias, particularly among subgroups such as English Language Learners (ELLs). Our SMD analysis revealed notable disparities across demographic groups, echoing previous research on the disproportionate impact of automated systems on linguistically diverse populations. While the current pipeline demonstrates strong overall performance, these disparities underscore the importance of ongoing bias mitigation strategies. However, we know that SMDs can be unreliable, especially for subgroups with smaller populations. Future work should explore methods such as regression-based adjustments, fairness-aware training techniques, or reinforcement learning approaches that explicitly account for subgroup characteristics.

\bibliography{references}

\end{document}